\documentclass[5p,times]{elsarticle}

\usepackage{lineno,hyperref}
\modulolinenumbers[5]

\journal{Journal of NeuroComputing}

\usepackage{amsmath}
\usepackage{amsfonts}
\usepackage{booktabs}
\usepackage{graphicx}

\begin{document}

\begin{frontmatter}

\title{Representation Learning with Deconvolution for Multivariate Time Series Classification and Visualization} 

\author[zzaddress]{Wei Song\corref{mycorrespondingauthor}\fnref{fn1}}
\author[mymainaddress]{Zhiguang Wang\fnref{fn1}}
\cortext[mycorrespondingauthor]{Corresponding author: Wei Song.}
\fntext[fn1]{These authors contribute equally to this work.}

\ead{iewsong@zzu.edu.cn}


\author[mymainaddress]{Lu Liu}
\author[northaddress]{Fan Zhang}
\author[zzaddress2]{Junxiao Xue}
\author[zzaddress]{Yangdong Ye}
\author[zzaddress]{Ming Fan}
\author[zzaddress2]{MingLiang Xu}
\address[zzaddress]{School of Information Engineering, Zhengzhou University, Zhengzhou, 450001, China}
\address[mymainaddress]{Department of Computer Science and Electric Engineering, University of Maryland Baltimore County, Baltimore, Maryland, 21228, United States}
\address[northaddress]{School of Information Engineering, North China University of Water Resources and Electric Power, Zhengzhou, 450046, China}
\address[zzaddress2]{Center for Interdisciplinary Information Science Research, Zhengzhou University, Zhengzhou, 450001, China}

\begin{abstract}
	We propose a new model based on the deconvolutional networks and SAX discretization to learn the representation for multivariate time series. Deconvolutional networks fully exploit the advantage the powerful expressiveness of deep neural networks in the manner of unsupervised learning. We design a network structure specifically to capture the cross-channel correlation with deconvolution, forcing the pooling operation to perform the dimension reduction along each position in the individual channel. Discretization based on Symbolic Aggregate Approximation is applied on the feature vectors to further extract the bag of features. We show how this representation and bag of features helps on classification. A full comparison with the sequence distance based approach is provided to demonstrate the effectiveness of our approach on the standard datasets. We further build the Markov matrix from the discretized representation from the deconvolution to visualize the time series as complex networks, which show more class-specific statistical properties and clear structures with respect to different labels.
\end{abstract}


\end{frontmatter}


\section{Introduction}



Sensors are now becoming cheaper and more prevalent in recent years to motivate the broad usage of
large amount of time series data. For example, Non-invasive, continuous, high resolution vital signs data, such as Electrocardiography (ECG) and Photoplethysmograph (PPG), are commonly used in hospital settings for better monitoring of patient outcomes to optimize early care. Industrial time series help the engineers to predict and get early preparation of the potential failure. We formulate these tasks as regular multivariate time series classification/learning problem. Compared to the univariate time series, multivariate time series is more ubiquitous, hence providing more patterns and insight of the underlying phenomena to help improve the classification
performance. Therefore, multivariate time series classification is becoming more and more important in a broad range of applications, such as industrial inspection and clinical monitoring.

Multivariate time series data is not only characterized by individual attributes, but also by the relationships between the attributes \cite{banko2012correlation}. Such information is not captured by the similarity between the individual sequences \cite{weng2008classification}. To deal with the classification problem on multivariate time series, several similarity measurements including Edit distance with Real Penalty (ERP) and Time Warping Edit Distance (TWED) are summarized and tested on several benchmark dataset \cite{Lin2012pattern}. Recently, a symbolic representation for multivariate time series classification (SMTS) is proposed. Mining core feature for early classification (MCFEC) along the sequence is proposed to capture the shapelets in each channel independently \cite{he2015early}. SMTS builds a tree learner with two ensembles to learn the segmentations and a high-dimensional codebook \cite{baydogan2014learning}.  While these methods provide new perspective to handle
multivariate data, some are time consuming (e.g. SMTS), some are effective but cannot address the curse of dimensionality (distance on raw data). 

Inspired by recent advances in feature learning for image classification, several feature-based approaches are proposed like \cite{baydogan2014learning,wang2015pooling}. Compared with those sequence-distance based approaches, the feature-based approaches skip the tricky hand-crafted features as they learn a hierarchical feature representation from raw data automatically. However, the feature learning approach are only limited on the scenario of supervised learning and few comparison towards distance-based learning approaches (like \cite{zheng2014time}). The method described in \cite{wang2015pooling} is simple but not fully automated, instead they still need to design the weighting scheme manually.

Our work provides a new perspective to learn the hidden representations with deconvolutional networks (in the self-supervised learning way), hence to fully exploit the unlabeled data especially when the data is large. We design the network structure to capture the cross-channel correlation with convolutions, forcing the pooling operation to perform the dimension reduction along each position of the individual channel. Inspired by the discretization approaches like Symbolic Aggregate Approximation (SAX) with its variations \cite{ lin2003symbolic, sun2014improvement,wang2015pooling} and Markov matrix \cite{wang2015imaging}, we further show how this representation helps on classification and visualization tasks. A full comparison with the sequence distance based approach is provided to demonstrate the effectiveness of our approach.  

\section{Background and Related Work}
\subsection{Deep Neural Networks}
Since 2006, the techniques developed from deep neural networks (or, deep learning) have greatly impacted natural language processing, speech recognition and  computer vision research \cite{bengio2009learning,LiDeep2014}. One successful deep learning architecture used in computer vision is convolutional neural networks (CNN) \cite{lecun1998gradient}. CNNs
exploit translational invariance by extracting
features through receptive fields \cite{hubel1962receptive} and learning
with weight sharing, becoming the state-of-the-art approach in various
image recognition and computer vision tasks
\cite{krizhevsky2012imagenet}. 
Most exciting advance comes from the exploration of unsupervised learning algorithms for generative models, such as Deep Belief Networks (DBN) and Denoised Auto-encoders (DA) \cite{hinton2006fast,vincent2008extracting}.  Many deep generative models are developed based on energy-based model or auto-encoders. Temporal autoencoding is integrated with Restrict Boltzmann Machines (RBMs) to improve generative models \cite{hausler2013temporal}. A training strategy inspired by recent work on optimization-based learning is proposed to train complex neural networks for imputation tasks \cite{brakel2013training}. A generalized Denoised Auto-encoder extends the theoretical framework and is applied to Deep Generative Stochastic Networks (DGSN) \cite{bengio2013generalized,bengio2013deep}.

However, since unsupervised pretraining has been shown to improve performance in both fully supervised tasks and weakly supervised tasks
\cite{erhan2010does, grzegorczyk2016encouraging}, deconvolution and Topographic Independent
Component Analysis (TICA) are integrated as unsupervised pretraining
approaches to learn more diverse features with complex invariance
\cite{zeiler2010deconvolutional, ngiam2010tiled, wang2016efficient}. We use deconvolution to capture both the temporal and cross-channel correlation in the multivariate time series, rather than pretrain a supervised model.

\subsection{Discretization and Visualization for Time Series} 
Time Series discretization is broadly used in symbolic approximation based approach . Aligned Cluster Analysis (ACA) is introduced as an unsupervised method to cluster the temporal patterns of human motion data \cite{zhou2008aligned}. It is an extension of kernel k-means clustering but requires quite computational capacity. Persist is an unsupervised discretization methods to maximize the persistence measurement of each symbol \cite{morchen2006finding}.  Piecewise Aggregate Approximation (PAA) methods is proposed by Keogh \cite{keogh2001dimensionality} to reduce the dimensionality of time series, which is then upgraded to Symbolic Aggregate Approximation (SAX) \cite{lin2003symbolic}. In SAX, each aggregation value after PAA process is mapped into the equiprobable intervals based on standard normal distribution to produce a sequence of symbolic representations. Among these symbolic approaches, SAX method has become one of the de facto standard to discretize time series and is at the core of many effective classification algorithms.

The principal idea of SAX is to smooth the input time series using Piecewise Aggregation Approximation (PAA) and assign symbols to the PAA bins. The overall time series trend is extracted as a sequence of symbols. 

\begin{figure}[t]
	\centering
	\includegraphics[width=0.5\textwidth]{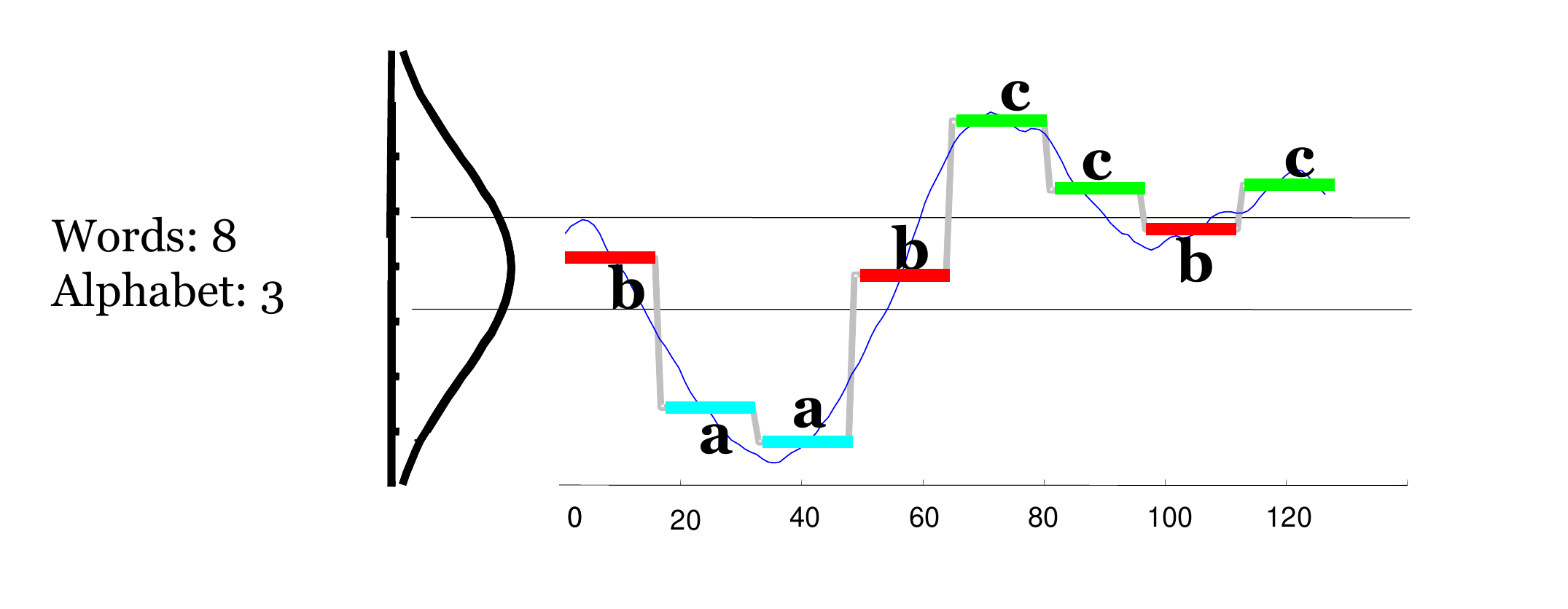}
	\caption {PAA and SAX word for the ECG data. The time series is partitioned into 8 segments. In each segment we compute means to map them to the equiprobable interval. After discretization by PAA and symbolization by SAX, we convert the time series into SAX word sequence $baabccbc$.}
	\label{fig:SAXdemo}
\end{figure}

The algorithm requires three parameters: window length $n$, number of symbols $w$ and alphabet size $a$. Different parameters lead to different representations of the time series. Given a normalized time series of length $L$, we first reduce the dimensionality by dividing it into $[L/n]$ non-overlapping sliding windows with skip size 1.  Each sliding window is partitioned into $w$ subwindows. Mean values are computed to reduce volume and smooth the noise. Then PAA values are mapped to a probability density function $\mathcal{N}(0,1)$, which is divided into several equiprobable segments. Letters starting from A to Z are assigned to each PAA values according to their corresponding segments (Figure \ref{fig:SAXdemo}).

In another hand, reformulating time series as visual clues has raised much attention in computer science and physics in which the discretization method plays an important role. The typical examples are that acoustic/speech data input is typically represented by Mel-frequency cepstral coefficients (MFCCs) or Perceptual Linear Prediction (PLP) to explicitly represent the temporal and frequency information. Researchers are trying to build different network structures from time series for visual inspection or designing distance measures. Recurrence Networks were proposed to analyze the structural properties of time series from complex systems \cite{donner2010recurrence,donner2011recurrence}. They build adjacency matrices from the predefined recurrence functions to interpret the time series as complex networks. Silva et al. extended the recurrence plot paradigm for time series classification using compression distance \cite{silva2013time}. Another way to build a weighted adjacency matrix is extracting transition dynamics from the first order Markov matrix \cite{campanharo2011duality}. Although these maps demonstrate distinct topological properties among different time series, it remains unclear how these topological properties relate to the original time series since they have no exact inverse operations. \cite{wang2015imaging} proposed an generalized Markovian encoding to map the complex correlations in the time series into images while preserving the temporal information as well. 

To give a intuition about how our learned feature is shaped, we simply build the Markov Matrix to visualize the topology of the formed complex networks as given by \cite{campanharo2011duality}.  

\section{Representation Learning Using Deconvolutional Networks}
Deconvolutional networks have the similar mathematical form with convolutional networks. The difference is, deconvolutional networks contain the 'inverse' operation of convolution and pooling for reconstruction.
 
\subsection{Deconvolution} 
\begin{figure}[t]
	\centering
	\includegraphics[width=0.5\textwidth]{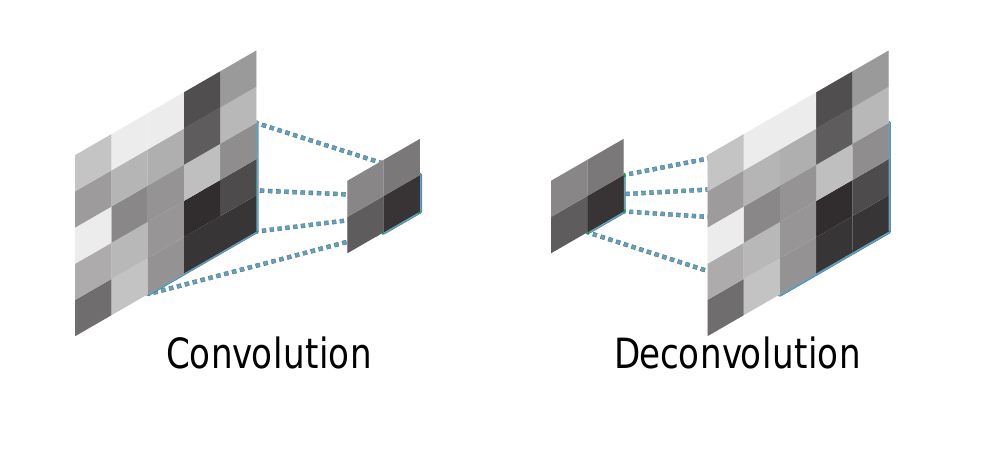}
	\caption{Illustration of the deconvolution operations.} 
	\label{fig:Deconvolution}
\end{figure}

Convolutional layers connect multiple input activations within a filter window to a single activation. In contrary, deconvolutional layers associate a single input activation with multiple outputs (Figure \ref{fig:Deconvolution}). The output of the deconvolutional layer is an enlarged and dense feature map. In practice, we crop the boundary of the enlarged feature map to keep the size of the output map identical to the one from the preceding unpooling layer.  

The learned filters in deconvolutional layers are actually matching the bases to reconstruct the same shape of the input, thus, similar to the convolution network, a hierarchical structure of deconvolutional layers are used to capture different level of shape details. The low level filters tend to capture detailed/fine-grained features while the filters in higher layers tends to capture more abstract features. Thus, the network directly takes specific shape information into account for multi-scale feature capturing, which is often ignored in other approaches based only on convolutional layers.

For the deconvolution operation, the feature maps are calculated as 
\begin{eqnarray}
H_{j} & = & \phi(Z_j) \nonumber \\ 
& = & \phi(\sum_{i}\delta(X_i) \otimes W_{ij} + b_j)
\end{eqnarray}

where $X_i$ represents the i-th element of input and $H_j$ denotes the j-th filter map after convolution and activation. The function padding function $\delta$ pads $X_i$ with zeros to keep the output size same with the input. After deconvolution, $H_j$ will be processed through a pooling layer. The reconstruction of $X_i$ is built based on $H_j$ in a reversed procedure of convolution. The reconstruction works in form of

\begin{eqnarray}
Y_{j} & = \phi(\sum_{i} \delta(\hat{H_i}) \otimes \hat{W_{ij}} + c_j)
\end{eqnarray}

$c_j$ is the bias term for the reconstruction $Y_j$. $\hat{H_i}$ is the feature map extracted by the unpooling layer. The gradient of each parameter in back propagation could be obtained and propagated to the preceding layers in an end-to-end manner until final convergence.

\subsection{Unpooling} 
\begin{figure}[t]
	\centering
	\includegraphics[width=0.5\textwidth]{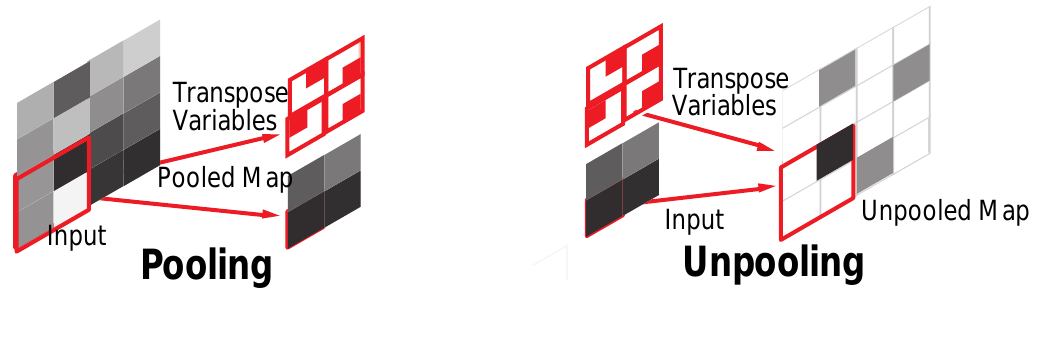}
	\caption{Illustration of the unpooling operations.} 
	\label{fig:Unpooling}
\end{figure} 

Pooling in convolution network abstracts activations in a receptive field with a single representative value to gain the robustness to noise and translation. Although it helps classification by retaining only robust activations in upper layers, spatial information within a receptive field is lost during pooling. Such information loss may be critical for precise feature learning that is required for reconstruction and classification. 

Unpooling layers in deconvolution network perform the reverse operation of pooling and reconstruct the original size of activations as illustrated in Figure \ref{fig:Unpooling}. To implement the unpooling operation, we record the locations of the maximum activations selected during pooling operation in the transposed variables, which are employed to place each activation back to its original pooled location. This unpooling strategy is particularly useful to reconstruct the structure of input object. Note that the output of an unpooling layer is an enlarged, yet sparse activation map, which might loss the expressiveness of the complex feature for reconstruction. To resolve the issue, the deconvolution layers is used after the unpooling operation to densify the sparse activations through convolution-like operations with multiple learned filters.

\subsection{Deconvolution for Multivariate Time Series}
In the proposed algorithm, the deconvolution network is a key component for precise feature learning on the multivariate time series data. Contrary to the simple usual deconvolution and pooling  both performed with square kernels, our algorithm generates feature maps using deep deconvolution network across the channel but pooling along each individual channel. The dense element-wise deconvolutional map is obtained by successive operations of unpooling, deconvolution, and rectification.

\begin{figure}[t]
	\centering
	\includegraphics[width=.5\textwidth]{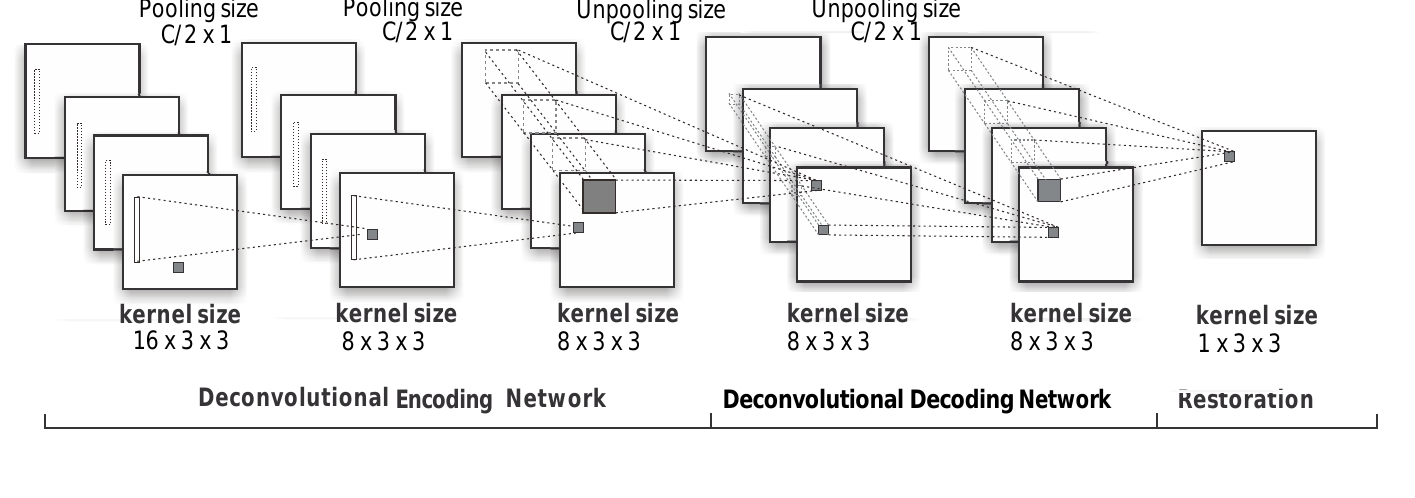}
	\caption{Demo of the complete network architecture for deep deconvolutional net on the multivariate time series.} 
	\label{fig:networkdemo}
\end{figure}

Figure \ref{fig:networkdemo} visualizes the example network structure layer by layer, which is helpful to understand internal operations of our deconvolution network. We can observe that deconvolution with multiple $3 \times 3$ filters are applied to capture both the temporal and cross-channel correlation. Lower layers tend to capture overall coarse configuration of the short term signals (e.g. location and frequency), while more complex patterns are discovered in higher layers. Note that pooling/unpooling layer and deconvolution play different roles for the construction of the learned features. Pooling/unpooling captures the significant information within a single channel by tracing each individual position with strong activations back to the signal space. As a result, it effectively reconstructs the detailed structure of the multivariate signals in finer resolutions. On the other hand, learned filters in deconvolutional layers tend to capture the generic generating shapes. Through deconvolution and tied weights, the activations closely related to the generating distribution along each signal and cross the channels are amplified while noisy activations from other regions are suppressed effectively. By the combination of unpooling and deconvolution, our network is able to generates accurate reconstruction of the multivariate time series.

\section{Visualization and Classification}
To visualize the learned representation to inspect and understand, we choose to discretize and convert the final encoding in the hidden layers of the deconvolutional networks to a Markov Matrix, hence visualizing them as complex networks \cite{campanharo2011duality}.  

\begin{figure}[t]
	\centering
	\includegraphics[width=.5\textwidth]{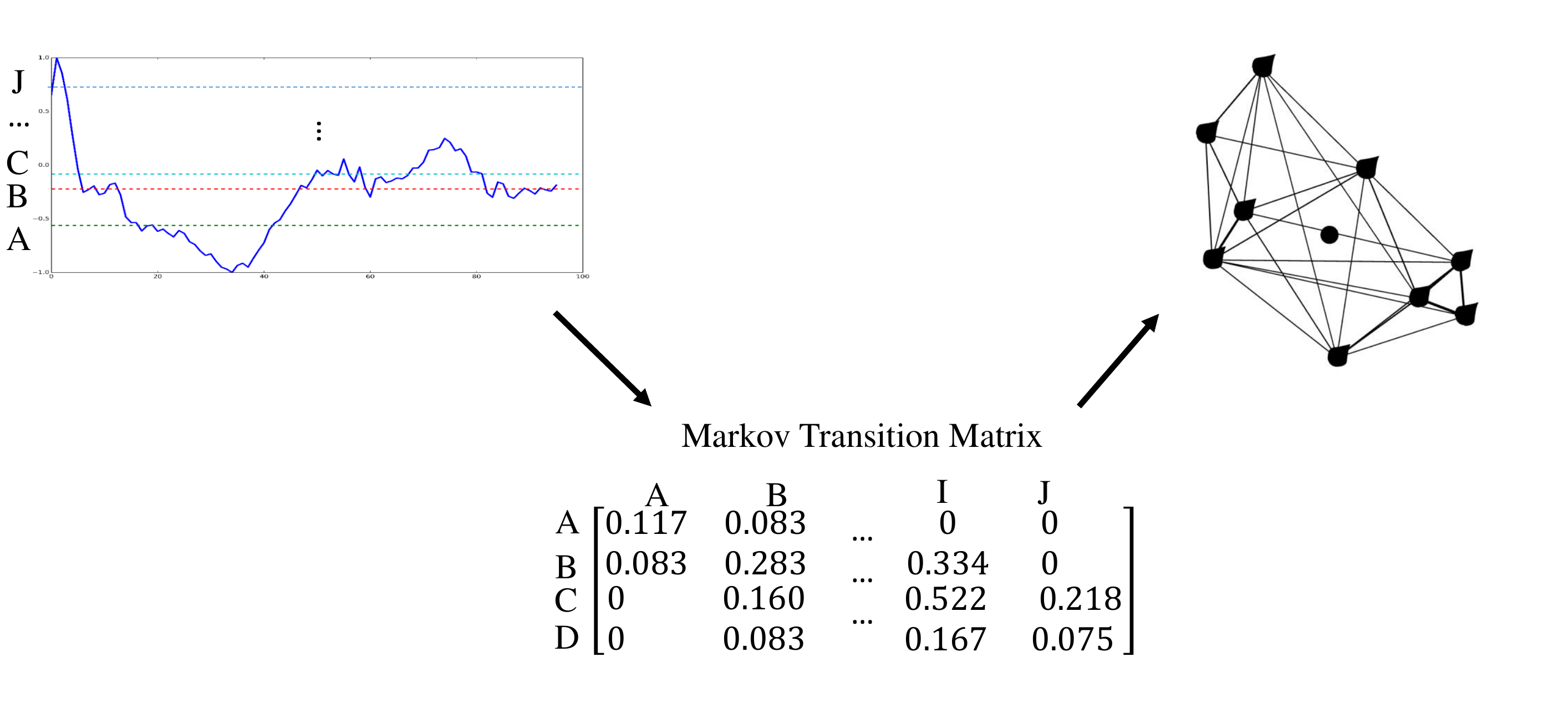}
	\caption{Illustration of the conversion from time series to complex networks for visualization}
	\label{fig:MarkovGraph}
\end{figure}

As in Figure \ref{fig:MarkovGraph}, a time series $X$ is split into $Q=10$ quantiles, each quantile $q_i$ is assigned to a node $n_i \in N$ in the corresponding network $G$. Then, nodes $n_i$ and $n_j$ are connected in the network with the arc where the weight $w_{ij}$ of the arc is given by the probability that a point in quantile $q_i$ is followed by a point in quantile $q_j$ . Repeated transitions between quantiles results in arcs in the network with larger weights, hence the connection is represented by thicker lines. Note that the discretization is originally based on quantile bins. As indicated in SAX methods that time series tends to follow the Gaussian distribution, we use Gaussian mapping to replace quantile bins for discretization. 

The deconvolution operation has a sliding window along time, which means the hidden representation should maintain a significant temporal component, thus be particularly within the application domain of SAX and bag-of-words approaches. The bag-of-words dictionary built from the SAX words is benefit from the invariance to locality. Compared with the raw vector-based representation, these feature bags improves the classification performance as it fits the temporal correlation while increase the expressiveness against noise and outliers. In our experiments, we use both the raw hidden vector and the bag of SAX words for classification.     

\section{Experiments and Results}   

This section first describes the settings and results of representation learning with deconvolution. Then, we analyze and evaluate the proposed representation in classification and visualization tasks.

We primarily use two standard datasets that are broadly appeared in the literature about multivariate time series \footnote{http://www.cs.cmu.edu/$\sim$bobski/}. The ECG dataset contains 200 samples with two channels, among which 133 samples are normal and 67 samples are abnormal.The length of a MTS sample is between 39 and 153. The wafer datasets contain 1194 samples. 1067 samples are normal and 127 samples are abnormal.The length of a sample is between 104 and 198. We preprocess each dataset by standardization and realigning all the signals with the maximum of the length. All missing values are filled by 0. Table \ref{tab:dataset} gives the statistics summary of each dataset.

\begin{table}[h]
	\centering
	\caption{Summary of the preprocessed datasets}
	\begin{tabular}{rrrrrr}
		\toprule
		& Channel & Length & Class & Training & Test \\
		\midrule
		Wafer & 6     & 199   & 2     & 896   & 298 \\
		ECG   & 2     & 153   & 2     & 100   & 100 \\
		\bottomrule
	\end{tabular}%
	\label{tab:dataset}%
\end{table}%

\begin{figure}[t]
	\centering
	\includegraphics[width=.5\textwidth]{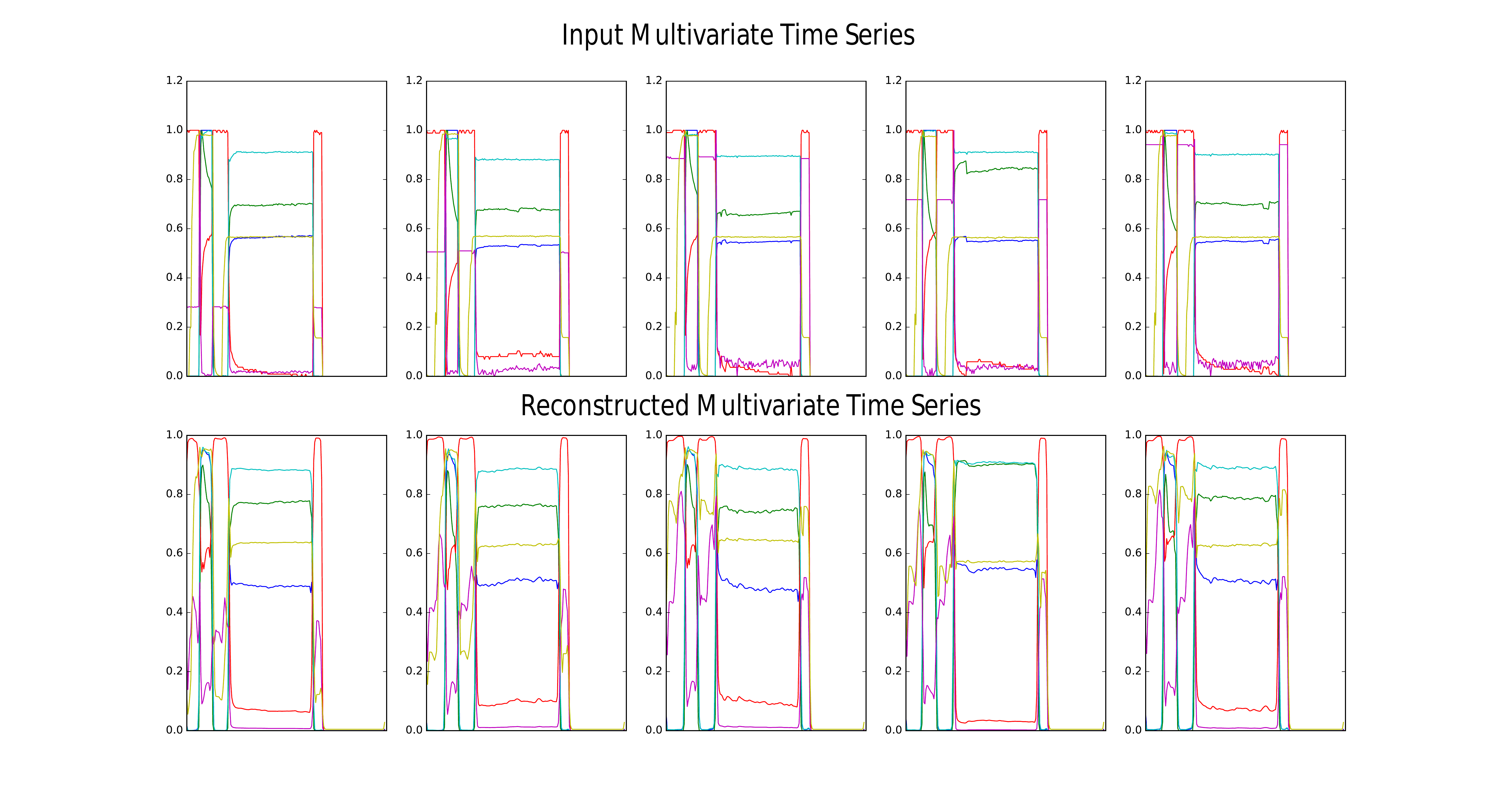}
	\caption{Input and reconstruction of our deconvolutional network on the 'wafer' dataset with 6 channels}
	\label{fig:recon_wafer}
\end{figure}

\begin{figure}[h!]
	\centering
	\includegraphics[width=.5\textwidth]{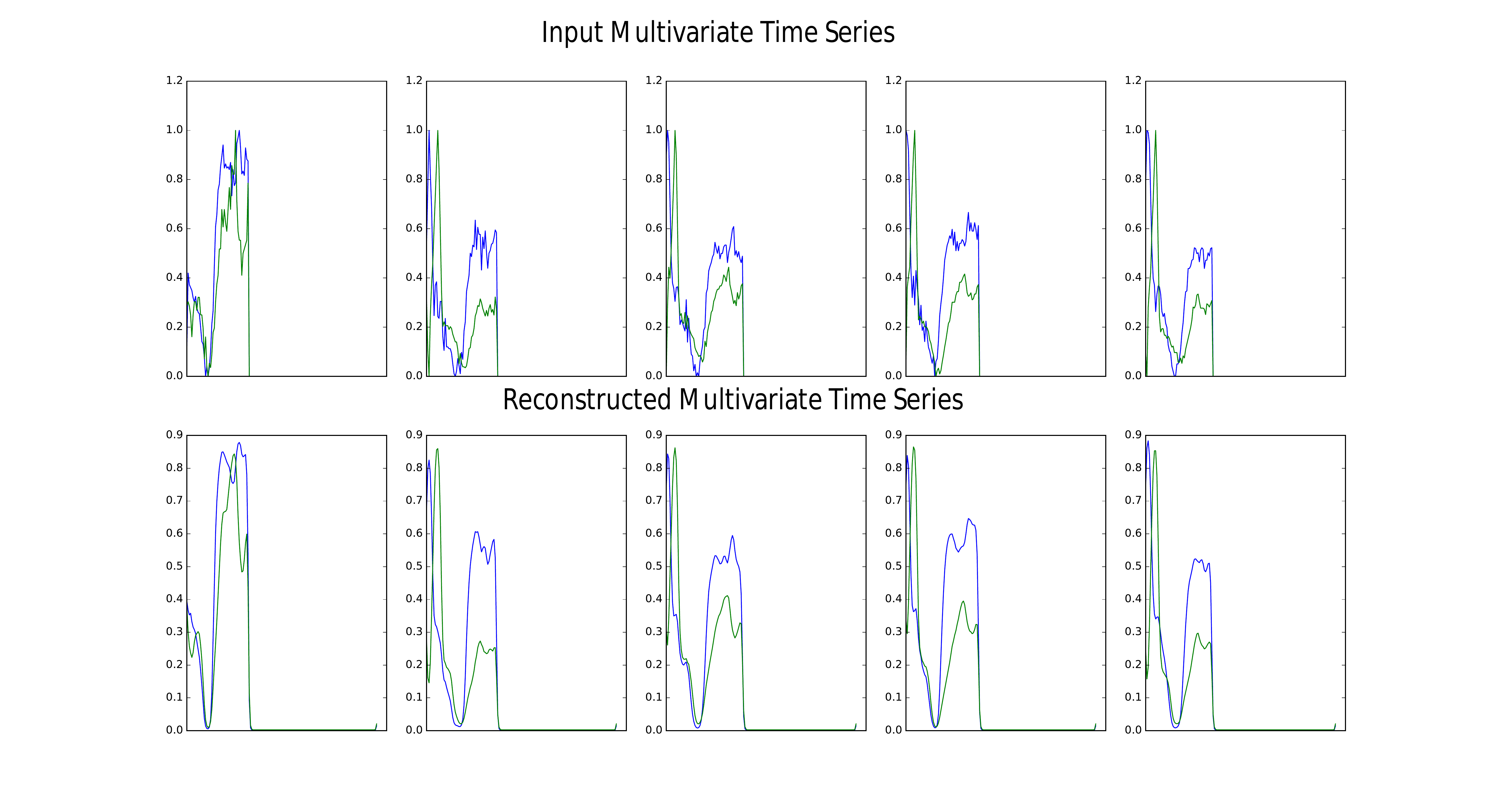}
	\caption{Input and reconstruction of our deconvolutional network on the 'ECG' dataset with 2 channels} 
	\label{fig:recon_ECG}
\end{figure}

\subsection{Representation Learning with Deconvolution}
Figure \ref{fig:networkdemo} summarizes the detailed configuration of the proposed network. Our network has symmetrical configuration of convolution and deconvolution network centered around the output of the 2nd Convolutional layer. The input and output layers correspond to input signals and their corresponding reconstruction. We use ReLU as activation function. The network is trained by Adadelta with learning rate $0.1$ and $\rho=0.95$ \footnote{Codes are available at https://github.com/cauchyturing/Deconv\_SAX}.  

\begin{figure}[t]
	\centering
	\includegraphics[width=.5\textwidth]{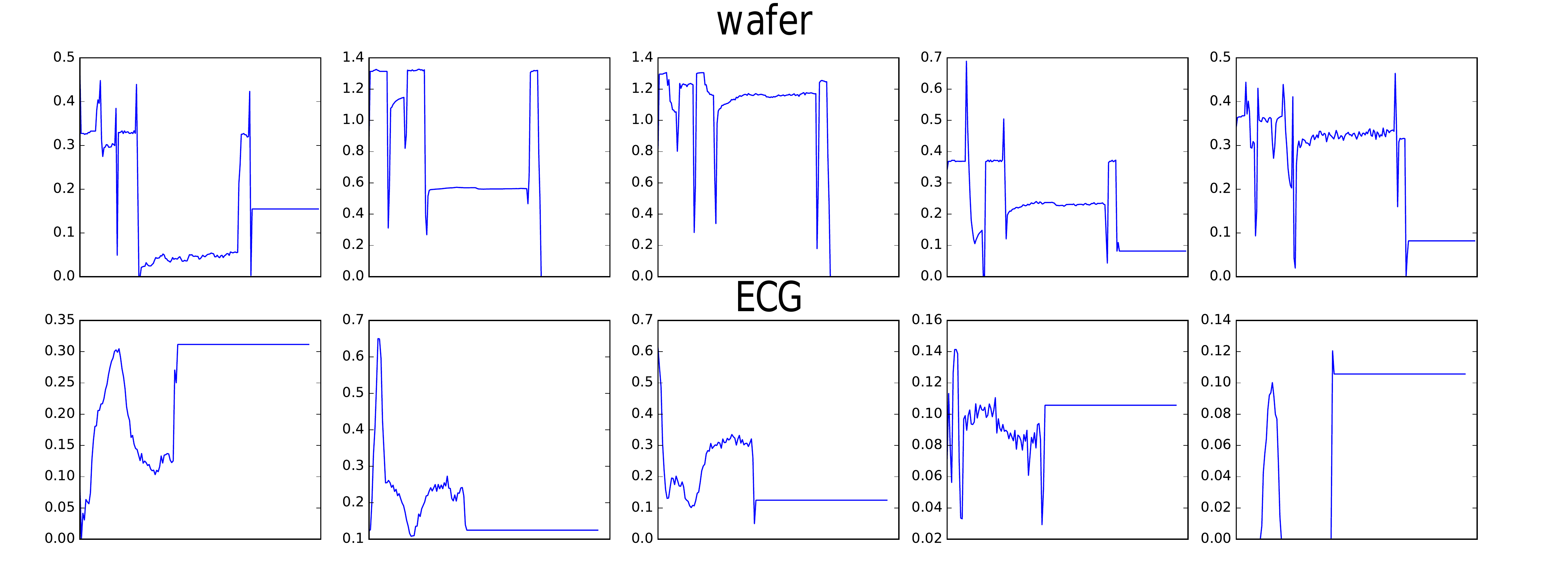}
	\caption{5 learned representation encoding in the deconvolutional networks.} 
	\label{fig:featuremap}
\end{figure}

Figure \ref{fig:recon_wafer} and \ref{fig:recon_ECG} show the reconstructions by our deconvolutional networks. While the filters trained by the deconvolution captures both the temporal and cross-channel information, combination of unpooling and deconvolution, our network is able to generates accurate reconstruction of the multivariate time series, which guarantees the expressiveness of the learned representations. As shown in Figure \ref{fig:featuremap}, After filtering by the deconvolution and pooling/unpooling, the final encoding of each map learned different representation independently. Diverse local patterns (shapes) of time series are captured automatically. Through the deconvolution, the filters determine the importance of each feature by considering both the single channel and cross channel information. 

\subsection{Classification}
For classification, we feed both the learned representation vector and the bag of SAX words into a linear SVM. Note that we only use training data to train the representation with deconvolutional networks, then generate the test representation using the well trained model with a single forward pass on the test set. The parameters of SAX, window length $n$, number of symbols $w$ and alphabet size $a$ is selected using Leave-One-Out cross validation in the training set with Bayesian optimization \cite{snoek2012practical}.

After discretization and symbolization, bag of words dictionary are built by a sliding window of length $n$ and convert each subsequence into $w$ SAX words.  Bag of words catch the features shared in the same structure among different instance and regardless of where they occur. The discretized features are built based on bag of words histogram of the word counts.

We compared our model with several best methods for multivariate time series classification in recent literatures including Dynamic Time Warping (DTW), Edit Distance on Real sequence (EDR), Edit distance with Real Penalty (ERP)\cite{Lin2012pattern}, STMS \cite{baydogan2014learning} and MCFEC \cite{he2015early} and Pooling SAX \cite{wang2015pooling}. 

\begin{table}[h]
	\centering
	\caption{Test error rate on the standard dataset}
	\begin{tabular}{rrr}
		\toprule
		& ECG   & Wafer \\
		\midrule
		Pooling SAX & 0.16  & \textbf{0.02} \\
		SMTS  & 0.182 & 0.035 \\
		MCFEC & 0.22  & 0.03 \\
		Euclidean $^\ast$  & 0.1778 & 0.0833 \\
		DTW(full) $^\ast$ & 0.1889 & 0.0909 \\
		DTW(window) $^\ast$ & 0.1722 & 0.0656 \\
		EDR  $^\ast$ & 0.2   & 0.3131 \\
		ERP  $^\ast$ & 0.1944 & 0.0556 \\
		\midrule
		Deconv (vector) & \textbf{0.13}  & 0.035 \\
		Deconv (SAX) & \textbf{0.13}  & \textbf{0.02} \\
		\bottomrule
	\end{tabular}%
	\label{tab:clf}%
\end{table}%

\begin{table*}[t!]
	\centering
	\caption{Summary statistics of the complex networks}
	\begin{tabular}{rrrrr}
		\toprule
		& Avg. Degree  & Modularity & Pagerank & Avg. Path Length \\
		\midrule
		ECG normal & 7.0966 & 0.5718 & 16.8  & 2.7802 \\
		ECG abnormal & 8.7202 & 0.5948 & 18.4  & 2.6384 \\
		P value  & \textbf{0.003} & 0.0124 & \textbf{0.0054 }& \textbf{0.0024} \\
		\midrule
		wafer normal & 8.7866 & 0.5058 & 12    & 2.0878 \\
		wafer abnormal & 8.5734 & 0.5098 & 12.2  & 2.0684 \\
		P value   & 0.2989 & 0.1034 & 0.3786 & \textbf{0.0012} \\
		\bottomrule
	\end{tabular}%
	\label{tab:netstats}%
\end{table*}%

Table \ref{tab:clf} summarizes the classification results\footnote{The results with $^ast$ are reported as the error rate of 10-fold cross validation on the whole datasets (train + test).}. Our model outperform all other approaches. Even wafer dataset has 6 channels, our approach is still able to capture the precise information through deconvolution to improve the classification performance. Because the datasets are small, supervised deep learning model tends to overfit the label, but in our unsupervised feature learning framework, the model takes advantage of the great expressiveness of the neural networks from the large number of weights to build precise feature set. These precise features provide a more optimal space for classification. 

Another comparison is performed between the vector and discretized bag-of-words representation from the deconvolutional networks (Table \ref{tab:vec_sax}). Although discretization by SAX introduces more hyperparameters, both cross validation and test error rate are better than the feature vector.  As we analyzed before, the deconvolutional networks learn the representation which preserve the high order abstract temporal information. Through SAX and bag of words, these information is enhanced particularly for classification in the supervised way. Noise and outliers that are less useful for classification are removed, while the constraint of the dependency on temporal locality is weakened by bag of words. Thus, the bag of SAX  words show advantage against the raw vector feature. The Bayesian optimization greatly facilitate the searching process on the hyperparameter and converge fast.       

\begin{table}[b]
	\centering
	\caption{Train and test error rates of the vector/discretized representation from the deconvolutional networks}
	\begin{tabular}{rrr|rr}
		\toprule
		& \multicolumn{2}{c}{SAX} & \multicolumn{2}{c}{Vector} \\
		\midrule
		& CV Train & Test  & CV Train & Test \\
		Wafer & 0.007 & 0.2   & 0.011 & 0.035 \\
		ECG   & 0.12  & 0.13  & 0.13  & 0.14 \\
		\bottomrule
	\end{tabular}%
	\label{tab:vec_sax}%
\end{table}%

\subsection{Visualization and Statistical Analysis}
Figure \ref{fig:featuremap} has shown how the vector representation is like as time series. To fully understand the representation learned through deconvolution and the effect of discretization through SAX, we flatten and discretize each feature map (which is feed in the classifier as input) and visualize them as complex networks to further inspect other statistical properties.

\begin{figure*}[t]
	\centering
	\includegraphics[width=.5\textwidth]{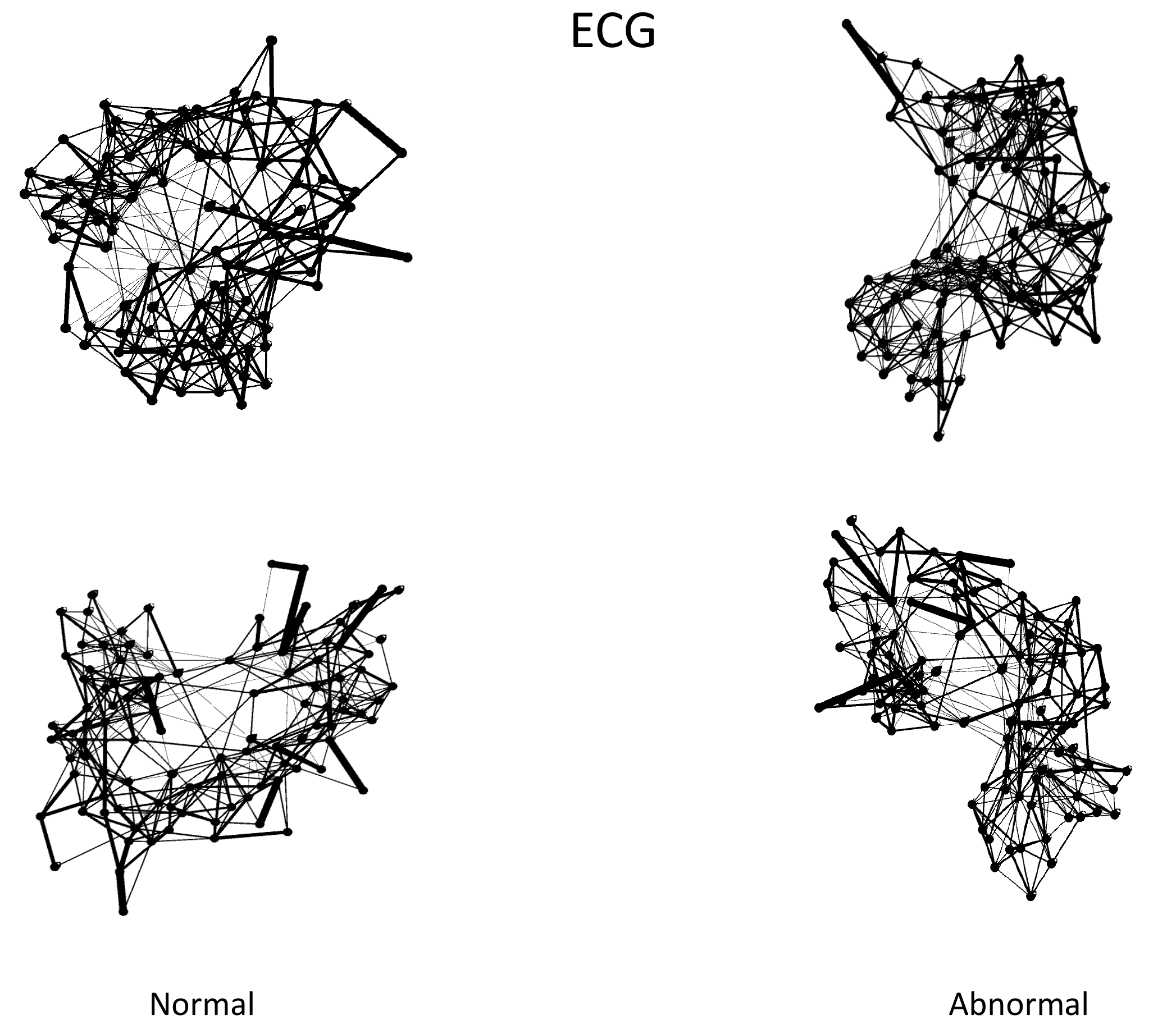}
	\caption{Visualization of the complex network generated from the discretized deconvolutional features on the ECG dataset.} 
	\label{fig:visecg}
\end{figure*}

\begin{figure*}[t]
	\centering
	\includegraphics[width=.5\textwidth]{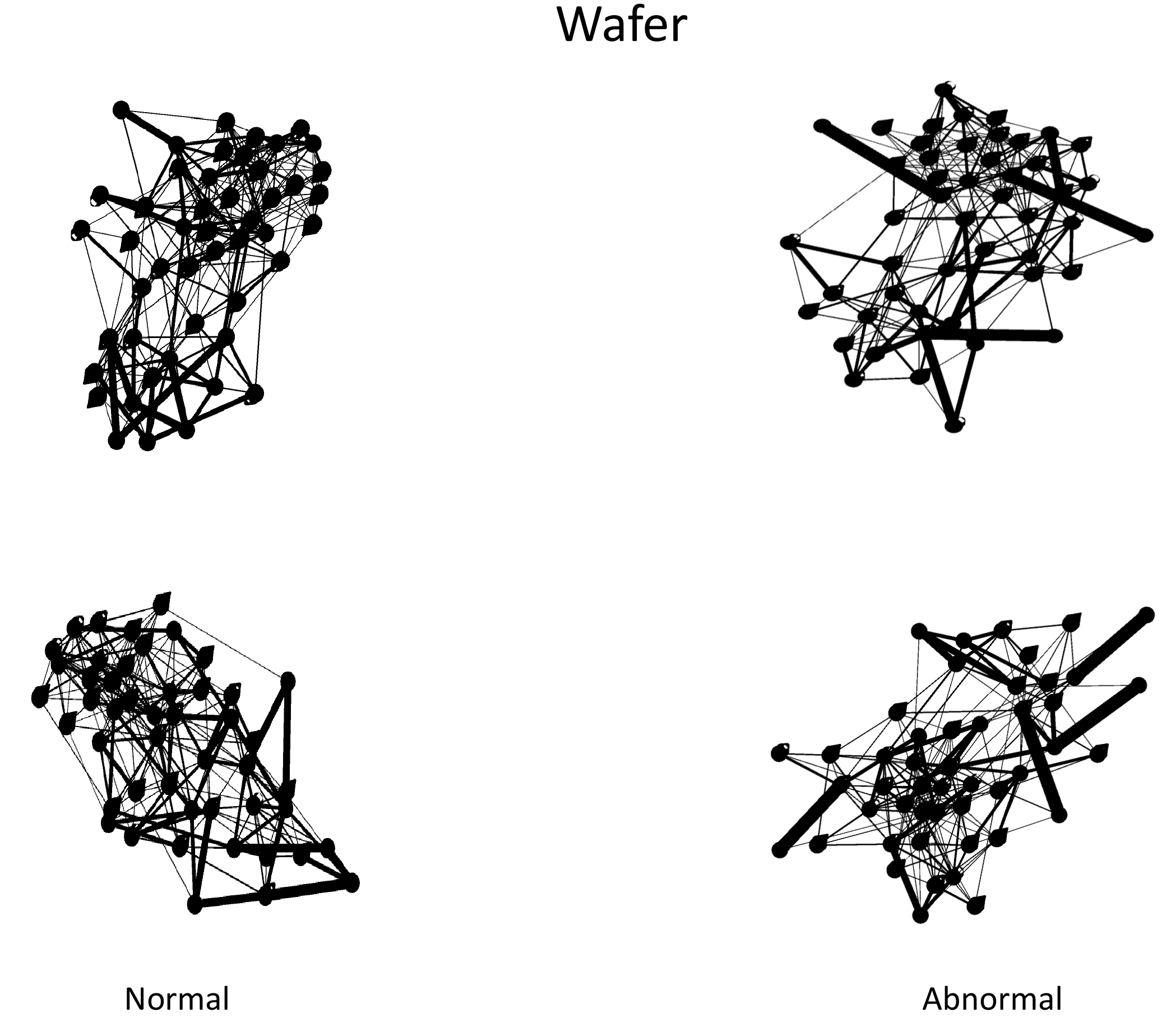}
	\caption{Visualization of the complex network generated from the discretized deconvolutional features on the Wafer dataset.} 
	\label{fig:viswafer}
\end{figure*}

As for the number of discretization bins (or the alphabet size in our SAX settings), we set $Q=60$ for the ECG dataset and $Q=120$ for the wafer dataset. We use the hierarchical force-directed algorithm as the network layout \cite{hu2005efficient}. As shown in Figure \ref{fig:visecg} and \ref{fig:viswafer}, the demo on the both dataset show different network structures. For ECG, The normal sample tends to have round-shaped layout while the abnormal sample always has a narrow and winded structure. As for wafer, normal sample is shown as a regular closed-form shape, but the structure of the abnormal sample is open while thicker edges piercing through the border.    

Table \ref{tab:netstats} summarizes four statistics of all the complex networks generated from the deconvolutional representations: average degree, modularity class \cite{blondel2008fast}, Pagerank index \cite{langville2011google} and average path length. Note that Pagerank index here denotes the largest value in its propagation distribution. For the ECG dataset, all statistics are significantly different between the graphs with different labels under the rejection threshold of 0.05. However, for the Wafer dataset, only average path length show significant difference between two labels. We think the reason is that thicker edges are appearing around the network structures, which indicates the number of edge and their weights are highly skewed. This topological structure would not effect other statistics, but would be reflected in the path degree.

\section{Conclusion and Future Work}
We propose a new model based on the deconvolutional networks and SAX discretization to learn the representation for multivariate time series. Deconvolutional networks fully exploit the advantage the powerful expressiveness of deep neural networks in the manner of unsupervised learning. We design a network structure specifically to capture the cross-channel correlation with deconvolution, forcing the pooling operation to perform the dimension reduction along each position in the individual channel. SAX discretization is applied on the feature vectors to further extract the bag of features.  We show how this representation and bag of features helps on classification. A full comparison with the sequence distance based approach is provided to demonstrate the effectiveness of our approach. We further build the Markov matrix from the discretized representation to visualize the time series as complex networks, which show more statistical properties and clear class-specific structures with respect to different labels.

As future work, we suppose to integrate grammar induction approach on the deconvolutional SAX words to further infer the semantics of multivariate time series. We are also interested in designing advanced intelligent interfaces to enable the interaction from human to inspect, understand and guide the feature learning and semantic inference for mechanical multivariate signals. 

\section{Acknowledgment}
This work is supported by the Natural Science Foundation of China (NSFC) under Grant No. 61472370, 61170223;The National Key Technology Research and Development Program of China under Grant No. 2013BAH23F01; The Education Department of Henan Province under Grant No.13A520453; The Department of Science \& Technology of Henan Province under Grant No.142300410229.

\bibliography{mybibfile}

\end{document}